# Polynomial Neural Networks Learnt to Classify EEG Signals

Vitaly Schetinin

TheorieLabor, Friedrich-Schiller-Universität Jena, Germany

Vitaly.Schetinin@uni-jena.de, http://nnlab.tripod.com

*Abstract* – *A neural network based technique is presented, which is able to successfully extract polynomial classification rules from labeled electroencephalogram (EEG) signals. To represent the classification rules in an analytical form, we use the polynomial neural networks trained by a modified Group Method of Data Handling (GMDH). The classification rules were extracted from clinical EEG data that were recorded from an Alzheimer patient and the sudden death risk patients. The third data is EEG recordings that include the normal and artifact segments. These EEG data were visually identified by medical experts. The extracted polynomial rules verified on the testing EEG data allow to correctly classify 72% of the risk group patients and 96.5% of the segments. These rules performs slightly better than standard feed-forward neural networks.*

## I. INTRODUCTION

Neural network based techniques have been applied to electroencephalogram (EEG) signals to classify mental tasks [1] and provide clinical interpretation [2, 3, 4, 5]. In general, the EEGs are non-stationary signals that vary from one patient to the other in a large range of amplitudes and frequency bands. Moreover, EEGs are distorted by artifacts and noise during recording. For these reasons, medical experts cannot objectively interpret the EEG recordings.

To learn classification rules from EEG data, it is used the unsupervised as well as the supervised learning. The first uses Kohonen's neural networks, and the second feed-forward networks trained by back-propagation algorithm.

To improve medical diagnostics, the authors [3] suggested to identify the type of the EEG corruptions. The features that characterize the ocular and muscle artifacts include spectral shapes. The extraction of these features are carried out by using techniques such as a parametric modeling and a cross correlation. Then three multi-layer neural networks subsequently use these features to perform blink, eye movement and muscle detection, respectively. Each detector was trained on real data with the artifacts. In the final stage of development the classification is enhanced by incorporating heuristic criteria such as a spatial distribution of the electrodes on the scalp. These heuristics are implemented as a set of rules in a rule-based system. The performance of the system was tested with 60 segments of the EEG test data the EEG experts pre-classified. Out of 1260 EEG segments, 94.5% were correctly classified.

Cluster analysis of EEG data has been also performed with a feed-forward neural network, which includes 72 input neurons and two output neurons with a sigmoid activation function [4]. A learning algorithm that has been suggested maximizes a Euclidean distance between output vectors belonging to different classes. Visualizing the outputs, the correlation between the clusters of the output data and the risk groups were found. The authors found that the sub-clusters, which belong to different classes, are strongly overlapped. In order to find a classification rule, they used a neural network with 24 input nodes corresponding to relevant features. Applying a pre-processing technique the authors finally selected 5 significant features from 72 initial features. They conclude firstly that the sub-clusters are overlapped because medical experts cannot certainly assign the patients to the risk groups. Secondly, the neural network classifiers must be trained on representative datasets consisting of well-defined patterns.

The system presented in [5], which consists of two artificial neural networks, was developed to assess dementia of Alzheimer type (DAT). The first network divides the DAT patients from non-DAT and the second estimates severity of the DAT. The EEGs were recorded via 15 electrodes, and then their power spectrums were calculated into 9 frequency bands. Additionally the relative power values were computed. The trained neural network correctly classified the DAT and non-DAT patients. The average error of severity score was 10%.

We can see that above fully connected neural networks learn to classify the EEGs well. However the corruptions of the EEGs by the artifacts still can produce misleading results. In particular, the blink, eye movement and muscle artifacts are similar to the brain activity characterizing by wave shape and frequencies.

Note also that the standard neural network techniques require to preset the suitable structures of the networks, that is, the users must properly define the input nodes as well as the number of hidden neurons of the networks. Moreover the users must apply the training methods that are able to prevent the neural networks from over-fitting which, as we know, reduces generalization ability of trained networks. In addition, classification rules that fully connected neural networks learnt from data can not be represented in a readable form due to a large number of connections between the input, hidden and output neurons.

Group Method of Data Handling (GMDH) based on a polynomial theory of complex systems has been invented by Ivakhnenko [6, 7, 8]. The GMDH does not require to preset the neural network structure and allows to comprehensively present a classification rule as a concise set of short-term polynomials.

To improve generalization ability of the GMDH-type networks, the authors [9, 10, 11] used a genetic inductive approach, which exploits a set of the short-term polynomials and a fitness function that penalize large network topology. However GMDH-type training algorithms are performed well if noise and distortions of the training data are distributed by a Gauss low. In a presence of many irrelevant features the training algorithms often over-fit the polynomial networks.

In this paper we consider two tasks: the fist is to develop a new training algorithm that is able to effectively prevent over-fitting the polynomial networks on real data and the second is to compare the standard and GMDH-type neural network techniques on real EEG data.

An algorithm we developed to train the polynomial networks is based on a projection method [12], which does not require to hypothesize the statistics of the noise and distortions. The polynomial rules we extracted from real EEG data allow to correctly classify 72% of the risk group patients and 96.5% of the EEG segments. These rules perform slightly better than the standard feed-forward neural networks

In section 2, we will describe some neural network classifiers which then will be compared on real data. In sections 3 we will describe in detail our training algorithm and then in section 4 we will describe the real EEG data we used to compare the performance of the trained neural networks. Finally in sections 5 and 6 we will discuss main results.

## II. NEURAL NETWORK CLASSIFIERS

A number of neural network approaches may be used to learn classification rules from EEG data. Here we describe firstly the feed-forward neural network classifiers trained by a standard back-propagation algorithm. Secondly we describe GMDH-type polynomial networks.

*II.A. Feed-Forward Neural Networks*

The feed-forward neural networks (FNNs) we used contain one hidden layer and one output neuron. A transfer function of neurons is a standard sigmoid $y = 1/(1 + \exp(-w_0 - \Sigma_i^m w_i x_i))$, where $x_i$ is a $i$-th input variable, $y$ is an output of neuron, $w_0$ and $w_i$ are a bias term and synaptic weights of neuron, respectively, and $m$ is the number of the input variables.

Note that neuron weights are initialized by random values. A structure of fully connected FNNs, as we know, is defined by users. The users must assign the input nodes and preset the number $h$ of the hidden neurons. Increasing step-by-step the numbers $m$ and $h$, the users may experimentally find out a FNN with the best classification accuracy. In our experiments we varied the numbers $m$ and $h$ and then trained each FNN by 100 times with different initial weights.

For training the FNNs, we exploited a fast Levenberg-Marquardt algorithm provided by the MATLAB. To prevent over-fitting the FNNs we used a standard earlier stopping technique that requires to divide the dataset into the training, testing and validating subsets.

*II.B. GMDH-Type Neural Networks*

In contrast to the FNNs, the GMDH-type neural networks do not need to preset their structures and may be comprehensively described by a concise set of polynomials [6, 7, 8]. The GMDH-type networks are the multi-layered ones consisted of the neurons whose transfer function $g$ is a short-term polynomial. For example, a linear polynomial is

$$y = g(u_1, u_2) := w_0 + w_1 u_1 + w_2 u_2, \quad (1)$$

where $u_1$ and $u_2$ are the variables, and $w_0$, $w_1$, $w_2$ are the neuron weights or the polynomial coefficients.

GMDH-type training algorithms are based on an evolutionary principle, which is performed as following. At the first layer $r = 1$, an algorithm, using all possible combinations by two from $m$ inputs, generates the first population of neuron-candidates. Since the neuron-candidates are fed by two different inputs, the number $L_1$ of the combinations, or a size of the population at the first layer, is equal to $C_m^2$.

In the first layer, the outputs of the neuron-candidates are $y_1^{(1)},…, y_{L1}^{(1)}$. Then an algorithm selects from this population of the neurons $F$ best ones, $F < L_1$. The selection of the best neurons is performed in accordance with a predefined fitness function whose value depends on the classification accuracy of the neurons-candidates. We can predefine such a criterion that its value is decreased when the classification accuracy of the neuron is increased.

In the second and next layers $r$, the size $L_r$ of the population defined by the number $F$, i.e., $L_r = C_F^2$. The generation and selection of the neurons are again performed. The new layers are created while the criterion value is decreased.

In Fig. 1 we depicted an example of the polynomial network consisting of 3 layers the GMDH algorithm grew for $m = 5$ inputs and $F = 4$.

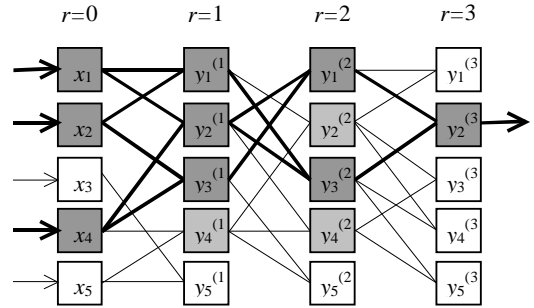

Fig. 1: An example of polynomial network

The neuron-candidates that were selected at each layers are depicted as grew boxes. A neuron $y_2^{(3)}$ that provides the best classification accuracy assigns to be an output neuron. A resulting polynomial network, as we can see, is 3 layer network consisting of 6 neurons and 3 input nodes. This network is described by a set of the following polynomials:

$$\begin{aligned}
y_2^{(3)} &= g_1(y_1^{(2)}, y_3^{(2)}), \\
y_1^{(2)} &= g_2(y_2^{(1)}, y_3^{(1)}), \\
y_3^{(2)} &= g_3(y_1^{(1)}, y_2^{(1)}), \\
y_1^{(1)} &= g_4(x_1, x_2), \\
y_2^{(1)} &= g_5(x_1, x_4), \\
y_3^{(1)} &= g_6(x_2, x_4),
\end{aligned} \quad (2)$$

where $g_1, …, g_6$ are the transfer function of the neurons.

To realize the selection procedure, a dataset are beforehand divided into two at least subsets. The first of them is used to train the neuron weights and the second to evaluate the classification accuracy of the neuron. Thus the value of the selection criterion depends on the behavior of the neuron on the examples that have not been included in the training subset. This kind of criteria called exterior allows to prevent GMDH-type networks from over-fitting [6, 7, 8].

The transfer function (1), the number $F$ of the neurons selected, as well as a selection criterion are predefined by users. Setting these parameters, the users may experimentally search of the best polynomial network. In the next section we will discuss in detail GMDH-type algorithms used for training the polynomial networks.

### III. GMDH-TYPE ALGORITHMS

In this section we firstly describe an idea of GMDH-type neural networks. As the networks are evolved during training, we would also say that the GMDH-type algorithm grows the networks. Secondly, we discuss the distortion of the real EEG data and describe a modified training algorithm.

*III.A. Training of GMDH-Type Networks*

Let $\mathbf{X}$ be a $n \times m$ matrix of input data that includes $n$ training examples presented by $m$ features, and $\mathbf{y}^o$ is a target vector: $\mathbf{y}^o = (y^o_1, \ldots, y^o_n)^T$, $y^o_i \in \{0, 1\}$. Denote a dataset $\mathbf{D} = (\mathbf{X}, \mathbf{y}^o)$.

Let a transfer function of neurons be a short-term polynomial of two variables $u_1$ and $u_2$. In nonlinear case, a transfer function $g$ includes the corresponding nonlinear terms, for example,

$$y = g(\mathbf{u}, \mathbf{w}) = w_0 + w_1 u_1 + w_2 u_2 + w_{12} u_1 u_2, \quad (3)$$

where $\mathbf{u} = (1, u_1, u_2, u_1 u_2)$ is a input vector, and $\mathbf{w} = (w_0, w_1, w_2, w_{12})$ are the polynomial coefficients or a weight vector.

At the first layer $r = 1$, the neurons are connected to the input nodes: the neuron inputs are given by pairs from $m$ variables $x_1, \ldots, x_m$:

$$\mathbf{u} = (1, x_{i1}, x_{i2}, x_{i1} x_{i2}), \quad i_1 \neq i_2 = 1, \ldots, m. \quad (4)$$

In the next layers $r = 2, 3, \ldots$, the inputs of the neurons are connected to the outputs $y_{i1}$ and $y_{i2}$ of the neurons from the previous ($r$ - 1) layer:

$$\mathbf{u} = (1, y_{i1}, y_{i2}, y_{i1} y_{i2}), \quad i_1 \neq i_2 = 1, \ldots, F. \quad (5)$$

Note that $F$ is the number of the best neurons selected from the previous layer. The users preset this number usually $F < 0.4 L_1$.

Thus, given the weight vector $\mathbf{w}$ and the $k$-th example for the input $\mathbf{u}^{(k)}$, we can calculate the output $y$ of the neuron:

$$y = g(\mathbf{u}^{(k)}, \mathbf{w}), \quad k = 1, \ldots, n. \quad (6)$$

For selecting $F$ best neurons, the GMDH uses the exterior criterion calculated on the unseen examples that have not been used for fitting the weights of the neurons. The unseen examples are reserved by dividing a dataset $\mathbf{D}$ into two non-intersecting subsets $\mathbf{D}_A = (\mathbf{X}_A, \mathbf{y}_A^o)$ and $\mathbf{D}_B = (\mathbf{X}_B, \mathbf{y}_B^o)$, named the training and examining datasets. The users define the sizes $n_A$ and $n_B$ of these subsets, usually, $n_A \approx n_B$, and $n_A + n_B = n$.

Let find out such a weight vector $\mathbf{w}^*$ that minimizes the sum square error $e$ of the neuron calculated on a subset $\mathbf{D}_A$

$$e = \Sigma_k (g(\mathbf{u}^{(k)}, \mathbf{w}) - y^o_k)^2, \quad k = 1, \ldots, n_A. \quad (7)$$

To find out a desirable minimum, the GMDH fits the neuron weights to a subset $\mathbf{D}_A$ by using a Least Square Method (LSM). Note a LSM may produce effective evaluations of weights if the distortions of the data are distributed by a Gauss law.

Assume that we have found a desirable weight vector $\mathbf{w}^*$ that minimizes the error (7) on a subset $\mathbf{D}_A$ for all $L_r$ neuron-candidates of the layer $r$. Then we can calculate the values $CR_i$ of the exterior criterion on a subset $\mathbf{D}_B$ that has not been used to fit the weights:

$$CR_i = \Sigma_k (g_i(\mathbf{u}^{(k)}, \mathbf{w}^*) - y^0_k)^2, \quad k = 1, \ldots, n_B, \; i = 1, \ldots, L_r. \quad (8)$$

We can see that the calculated value of $CR_i$ depends on the behavior of the $i$-th neuron-candidate on the unseen examples of the subset $\mathbf{D}_B$. Therefore we may expect the value of $CR$ calculated on entire set $\mathbf{D}$ will be high for the neurons with small generalization ability.

The values $CR_i$ calculated at the $r$-th layer are arranged in ascending order:

$$CR_{i1} \leq CR_{i2} \leq \ldots \leq CR_F \leq \ldots \leq CR_L, \quad (9)$$

so that the first $F$ neurons are the best.

For each layer $r$ it is defined a minimal value $CR_m$ corresponding to the best neuron, i.e., $CR^{(r)}_m = CR_{i1}$. The first $F$ best neurons are then used at the next $r + 1$ layer.

The outputs of $F$ selected neurons in accordance with (5) feed the neuron-candidates at the $r + 1$ layer. The training and selection of the neurons of this layer performed with the equations (7), (8) and (9) are repeated.

The value of $CR_m^r$ is step-by-step decreased while the number $r$ of the layers is increased and the network grows. Once, the value of $CR$ reaches to a minimal point and then starts to increase, for examples, see Fig. 4.

In Fig. 4 the value of $CR_m$ is decreased at the three layers of the polynomial network, i.e., $CR^{(1)}_m > CR^{(2)}_m > CR^{(3)}_m$. At the $r = 3$ layer the value of $CR_m$ becomes to be minimal. At the next $r = 4$ layer the value of $CR^{(4)}_m$ is increased, therefore in accordance with the exterior criterion the polynomial network has been over-fitted. Because a minimum of $CR$ was reached at the previous layer, we stop the training algorithm and conclude that a desirable polynomial network has been grown at the $r = 3$ layer.

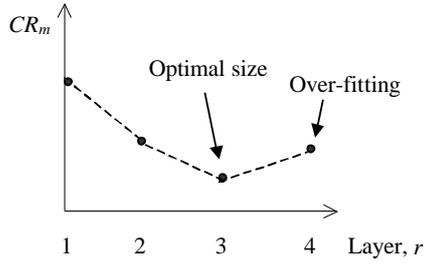

Fig. 2: The value of criterion $CR_m$ calculated for the grew networks

Recall that GMDH training algorithm assumes a Gauss distribution of the distortions in the data **X**. For real-world data with the unknown distribution function of distortions, below we describe a modified GMDH-type training algorithm.

*III.B. A Modified GMDH-Type Algorithm*

In many real-world tasks it is difficult to hypothesize a distribution function, including a Gauss one, suitable for training data. As an example, we depicted in Fig. 3 a distribution function for an EEG variable such as a real power of theta in C3.

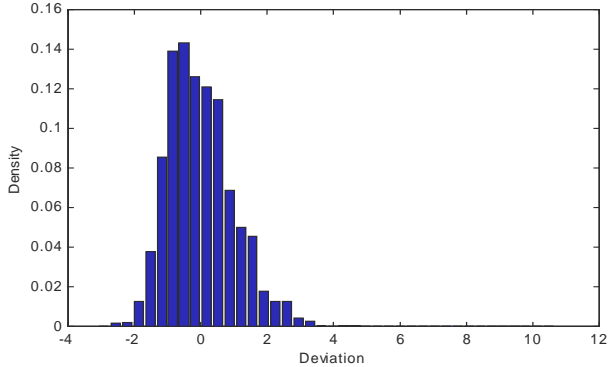

Fig. 3: The distribution function of EEG variable

We can see this distribution function is asymmetrical and its right tail much longer than the left. In addition, note that the distribution functions of all EEG variables are different. That is, it is required to hypothesize the distribution function for each variable individually.

To avoid above problems, we suggest to fit the neuron weights by using a recurrent algorithm based on a projection method [12]. Below we describe this algorithm in detail.

Let $\mathbf{u}_A$ and $\mathbf{u}_B$ be the $p$ input vectors that correspond to the training and examining examples from the subsets $\mathbf{X}_A$ and $\mathbf{X}_B$, respectively, where $p = 4$ defined by a given transfer function. Using these notations, we can describe the basic steps of our algorithm as following.

Initially $k = 0$, and an algorithm initiates a weight vector $\mathbf{w}^0$ by random values, for example, distributed by a normalized Gauss function. Then, at step $k = 1, 2, \ldots$, the algorithm calculates the $n_A \times 1$ error vector $\mathbf{\eta}_A^k$ on the training dataset $\mathbf{D}_A$. Its elements $\eta_{A\,i}^k$, $i = 1, \ldots, n_A$, are calculated as

$$\eta_A^k = g(\mathbf{u}_A, \mathbf{w}^{k-1}) - y_A^0, \quad y_A^0 \in \mathbf{Y}_A^0. \quad (10)$$

Correspondingly for a dataset $\mathbf{D}_B$, it is calculated the elements $\eta_{B\,i}^k$, $i = 1, \ldots, n_B$, of the $n_B \times 1$ error vector $\mathbf{\eta}_B^k$:

$$\eta_B^k = f(\mathbf{u}_B, \mathbf{w}^{k-1}) - y_B^0, \quad y_B^0 \in \mathbf{Y}_B^0. \quad (11)$$

Using the elements of the vector $\mathbf{\eta}_B^k$, we can now calculate a residual squared error (RSE) $E_B$ of the neuron on the validating dataset:

$$E_B(k) = (\Sigma_i \eta_{B\,i}^k)^{1/2}, \, i = 1, \ldots, n_B. \quad (12)$$

Now we can formulate a goal of the fitting algorithm that is to maximally reduce the error $E_B$ during a finite number $k$ of steps. This goal is achieved if the following inequality is satisfied at a step $k^*$:

$$E_B(k^*) \leq \varepsilon, \quad (13)$$

where $\varepsilon > 0$ is a given constant depending on the level of the noise in a dataset **X**.

If this inequality is not satisfied, then the current weight vector $\mathbf{w}^{k-1}$ is updated in accordance to the following learning rule

$$\mathbf{w}^k = \mathbf{w}^{k-1} - \chi \|\mathbf{U}_A\|^{-2} \mathbf{U}_A \mathbf{\eta}_A^{k-1}, \quad (14)$$

where $\chi$ is the learning rate, $\mathbf{U}_A = (\mathbf{u}_A^{(1)}, \ldots, \mathbf{u}_A^{(nA)})$ is the $p \times n_A$ matrix of the input data, $\|\mathbf{U}\|$ is an Euclidian norm of matrix **U**, $\|\mathbf{U}\| = (\Sigma_i^p u_i^{(1)2} + \ldots + \Sigma_i^p u_i^{(nA)2})^{1/2}$.

The rate $\chi$ has to be between 1 and 2, then the algorithm always yields the desired estimates of the weights with a given accuracy $\varepsilon$ during a finite number $k^*$ of steps (a proof see in [12]).

To explain this learning rule we depicted in Fig. 4 a space of two weight components $w_1$ and $w_2$ and a vector $\mathbf{w} = (w_1, w_2)$. Let us assume that in this space there is a desirable region $\mathbf{w}^*$ in which the inequality $E_B \leq \varepsilon$ is satisfied for each vector $\mathbf{w} \in \mathbf{w}^*$. However, at a step $k$ a vector $\mathbf{w}^k$ is beyond $\mathbf{w}^*$, $\mathbf{w}^k \notin \mathbf{w}^*$.

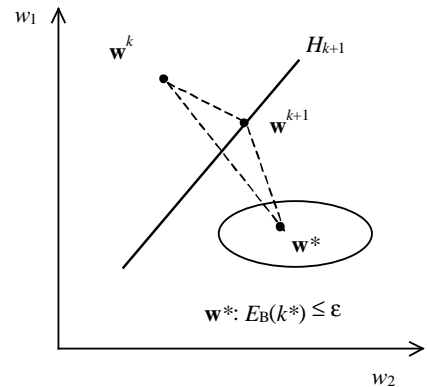

Fig. 4: A projection in a space of two weight components $w_1$ and $w_2$

Obviously, the value $E_B$ is proportional to a distance between a given vector $\mathbf{w}^k$ and a region $\mathbf{w}^*$. The next vector $\mathbf{w}^{k+1}$ updated by a rule (14) is an orthogonal projection of a vector $\mathbf{w}_k$ on a hyperplane $H_{k+1}$, which is located between a vector $\mathbf{w}^k$ and a region $\mathbf{w}^*$. An orthogonal projection on $H_{k+1}$ denoted as a new vector $\mathbf{w}^{k+1}$ is closer to $\mathbf{w}^*$ than a previous $\mathbf{w}^k$.

Since a new vector $\mathbf{w}^{k+1}$ is closer to a desirable region $\mathbf{w}^*$ than a previous $\mathbf{w}^k$, the RSE value $E_B(k+1) < E_B(k)$. Then we can conclude that $E_B(k) < E_B(k-1) < \ldots < E_B(0)$.

In our experiments we varied $\chi$ from 1.25 to 2.0 and obtained different learning curves, which depicted in Fig. 5. As we can see, the RSE is decreased with maximal speed if a learning rate $\chi = 2.0$.

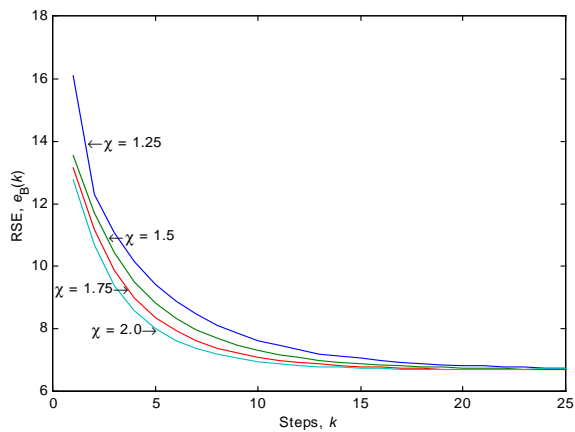

Fig. 5: The learning curves for $\chi = 1.25, 1.5, 1.75$ and $2.0$

In the cases when the level of the noise in the data set $\mathbf{X}$ is unknown, instead of a stopping rule (13) we can preset a constant $\Delta > 0$, which defines a minimal decrement of the RSE $E_B$, calculated at the step $k - 1$ and $k$, respectively. Then the goal of the training algorithm is achieved, if the RSE difference between step $k^*$ and $k^* - 1$ will be less than $\Delta$:

$$E_B(k^*-1) - E_B(k^*) < \Delta. \quad (15)$$

Thus, after $k^*$ steps, the algorithm provides a desired weight vector $\mathbf{w}^*$. Correspondingly, the value $CR_m$ of the fitness function is given by $CR_m = E_B(k^*)$.

In our experiments the best performance was obtained with $\chi = 1.9$ and $\Delta = 0.0015$. In this case the number $k^*$ usually did not exceed 30 steps, see, for example, at Fig. 5. Additionally we varied the configurations of the subsets $\mathbf{D}_A$ and $\mathbf{D}_B$ as well as the number $F$ between 40 and 80 in order to experimentally find out a minimal RSE.

## IV. EEG DATA

In this section we describe EEG data used in our experiments. The first EEG data available on [13] are the recordings made from Alzheimer and healthy patients. The second are the EEG recordings made from the sudden risk death patients and the third are the EEG recordings with the normal and artifact segments.

### IV.A. Alzheimer Data

We used two EEGs, which have been made from a healthy young person and an Alzheimer patient both with the opened eyes. The EEG data consist of 19 columns corresponding to 19 standard EEG channels. The sampling rate is 128 Hz, and total time is 8 seconds. Digital conversion of the measured signal was done with 8 bits. A band-pass filter (from 0.1 Hz to 30Hz) was used as well.

The medical viewer visually cleaned the EEG recordings by eliminating the segments, which contain the muscular artifacts. Spatial behavior is examined by accumulating data at various scalp locations. The standard 10-20 system for electrode placement was used that provide 19 simultaneous EEG measurements.

For each channel we calculated spectral power density value into four standard frequency bands: delta (0-3 Hz), theta (4-7 Hz), alpha (8-13 Hz) and beta (14-20 Hz). Thus the number of the input features is equal to $19*4 = 36$.

Following [1] we divided each 8-second recording into ½ second segments that overlap each ¼ second. Thus each segment includes 128 samples of the EEG. The first 4 seconds of the EEG recording we assigned to be a training set and the rest of 4 seconds to be the testing set. Therefore the training and testing sets we used consist of $2(4*2 - 1) = 14$ labeled segments.

### IV.B. Sudden Death Risk Data

The EEG recordings were made via two standard C3 and C4 channels with sample rate of 100 Hz. The EEG recordings of the patients were assigned by medical expert to three clinical groups: (i) is the healthy patients, (ii) is the patients with frequent apnea and (iii) is the patients with both frequent and pathological apnea.

Following [4], the power spectral densities were calculated during 10-second segments into 6 frequency bands: sub-delta (0-1.5 Hz), delta (1.5-3.5 Hz), theta (3.5-7.5 Hz), alpha (7.5-13.5 Hz), beta 1 (13.5-19.5 Hz), and beta 2 (19.5-20 Hz). In addition the relative and absolute power values as well as their amplitude variances were calculated. Then the EEG-viewer removed from the EEG data the artifact segments.

Thus we have 72 EEG features. These features have been then enhanced with 8 clinical features. Using these 80 features we have composed the training and testing sets both consisting of 43 patients belonging to the risk groups (i) and (iii).

To use the relevant features, we have performed a standard principle component analysis (PCA) the MATLAB provides. A PCA produced 3 principle components, which contribute 92% of the variance of the training dataset. The first and second components, which were calculated on the training dataset with PCA are depicted in Fig. 6.

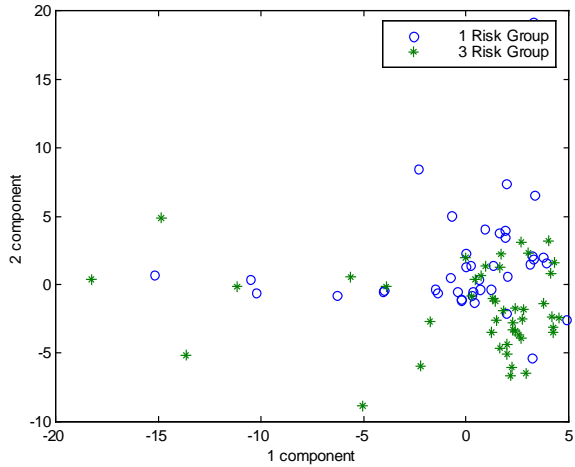

Fig. 6: The first and second components for (i) and (iii) risk groups

Fig. 6 shows that the (i) and (iii) risk group examples signed by the circles and the asterisks are substantially overlapped on the plane of the first two principle components. As we see, it is impossible to find a classification rule, which could correctly distinguish all labeled examples

*IV.C. EEG Artifact Data*

We used the EEG recordings made from two patients to train the neural-network to automatically recognize the artifact segments. The EEG expert using information from additional channels that reflect respiratory control, muscular activity, eye blinks has visually detected the artifact segments. Then the expert labeled the artifacts in these EEG recordings.

In this experiment the training and testing sets consisted of 1347 and 808 examples, respectively. Following to [4], each example was represented by 72 input features calculated on 10-second intervals. The training and testing sets included 88 and 71 artifacts, respectively. Fig. 7 depicts the distribution of the artifacts and the normal segments in a space of two principle components.

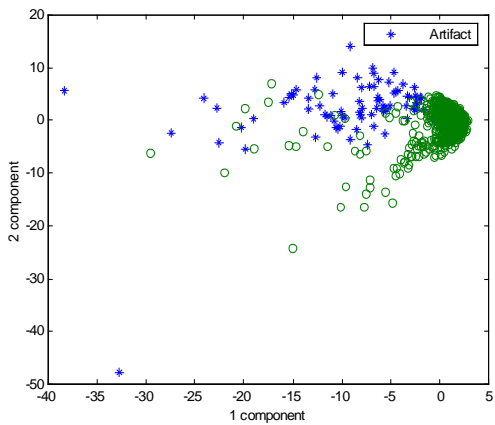

Fig. 7: The first two components for normal and artifact segments

In contrast to above case, we see that the normal and artifact examples, depicted respectively by the circles and the asterisks, are slightly overlapped. Therefore the classification of the EEG segments is expected to be more accurate than in the previous case.

## V. EXPERIMENTAL RESULTS

In this section we test the neural networks on real EEG data described in the previous section. Then we compare the accuracy of the standard FNNs and GMDH-type polynomial networks on these EEG data.

*V.A. Classification of Alzheimer EEG*

A standard PCA calculated on Alzheimer EEG data characterized by 76 features has produced 8 principle components. These features have been used as the input nodes of the FNN. The best of the FNNs trained by a back-propagation algorithm we found experimentally: this FNN includes 2 hidden neurons.

A polynomial neural network (PNN) trained by our algorithm has been learnt a classification rule that is described by a set of 3 polynomials:

$$y_1^{(1)} = 0.6965 + 0.3916\, x_{11} + 0.2484\, x_{69} - 0.2312\, x_{11} x_{69}, \quad (16)$$
$$y_1^{(2)} = 0.3863 + 0.5648\, y_1 + 0.5418\, x_{73} - 0.4847\, y_1 x_{73},$$
$$y_1^{(3)} = 0.1914 + 0.7763\, y_2 + 0.2378\, x_{76} - 0.2042\, y_2 x_{76}$$

where $x_{11}$ is delta in C11, $x_{69}$ is beta in C12, $x_{73}$ is beta in C16, and $x_{76}$ is beta in C19. All inputs were normalized.

As we can see this polynomial rule includes 4 features selected from $m = 76$ input variables. Fig. 8 depicts an appropriate structure of this classification rule. A PNN consists of 4 input nodes and 3 neurons whose transfer function is described by (3).

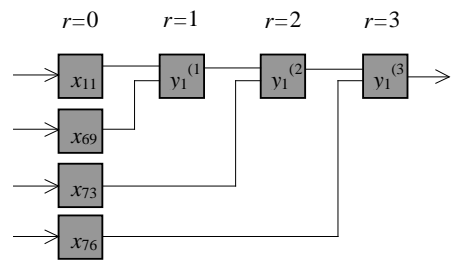

Fig. 8: A network classifying the EEGs of Alzheimer and healthy patients

Note that in this experiment the best PNN has been grown by the GMDH algorithm that selects one neuron at each layer, i.e. $F = 1$.

In general, in this case it is grown the simplest polynomial networks. At the first layer the algorithm forms the input vectors described by (4). However at the next layers, $r = 2, 3, \ldots$, the input vectors are $(1, y_1^{(r-1)}, x_i, y_1^{(r-1)} x_i)$, $i = 1, \ldots, m$.

The polynomial rule (16) may be used to classify EEGs as the following:

```
If the output y ≥ 0.5, then an EEG corre-
sponds to a healthy patient, otherwise to
an Alzheimer patient.
```

We also trained GMDH-type networks on these EEG data. However on the testing data all free neural networks correctly classified 13 EEG segments from 14.

*V.B. Classification of Sudden Death Risk*

For 80 features characterizing the sudden death risk data the PCA produced 3 components which contribute to the variance of the data 96%. Accordingly we trained the FNNs with 3 input nodes. The number hidden neurons we varied from 2 to 20 and found out that the best FNN contains 10 hidden neurons. This FNN has misclassified 15 testing examples from 43.

The standard GMDH-type neural network and the PNN have been grown with $m = 80$ inputs and $F = 40$ and $F = 60$, respectively. Note that in our experiments the GMDH with $F = 60$ produced the over-fitted polynomial networks.

In Table 1 we depicted the errors on the training and testing datasets for all three types of the trained neural networks. Note that here and further a PNN denoted a polynomial network trained by our algorithm.

Table 1: The errors on EEG data of sudden death risk patients

| Dataset | The number of errors | | |
|---------|------|------|-----|
|         | FNN  | GMDH | PNN |
| Train   | 8    | 3    | 3   |
| Test    | 15   | 13   | 12  |

From Table 1 we see that the trained PNN is performed slightly better than the best FNN and GMDH-type network. The PNN correctly classified 31 from 43 or 72,1% of the testing examples.

*V.C. Recognition of EEG Artifacts*

For the EEG artifact data described by 72 features, the PCA produced 8 principle components, which contribute to the variance of these data 92%. Correspondingly we trained the FNNs with 8 input nodes. The number of the hidden neurons we varied from 2 to 20. The best FNN trained with 10 hidden neurons has misclassified 24 testing examples.

The GMDH-type network and the PNN have been grown with $m = 72$ inputs and $F = 40$ and $F = 60$, respectively. Table 2 depicts the errors of the trained FNNs, GMDH-type neural networks and PNNs on the training and testing sets.

Note that in Table 2 we depicted the errors of the neural networks that were the best on the training set. For example, the best of 100 trained FNNs misclassified 27 training and 31 testing examples.

Table 2: The errors of neural networks on EEG artifact datasets

| Dataset | The number of errors | | |
|---------|------|------|-----|
|         | FNN  | GMDH | PNN |
| Train   | 27   | 35   | 30  |
| Test    | 31   | 33   | 28  |

We found that the best PNN that made 30 errors on the training set misclassified 28 testing examples. Appropriate classification rule that the PNN learnt from training data is described by the following set of 11 polynomials:

$$y_1^{(1)} = 0.9049 - 0.1707 x_5 - 0.1616 x_{57} + 0.0339 x_5 x_{57}, \quad (17)$$
$$y_2^{(1)} = 0.9023 - 0.2128 x_5 - 0.1389 x_{28} + 0.0438 x_5 x_{28},$$
$$y_3^{(1)} = 0.9268 - 0.1828 x_6 - 0.1195 x_{62} + 0.0233 x_6 x_{62},$$
$$y_4^{(1)} = 0.9323 - 0.2057 x_6 - 0.0461 x_{21} + 0.0246 x_6 x_{21},$$
$$y_5^{(1)} = 0.9247 - 0.1822 x_5 - 0.0951 x_{55} + 0.0196 x_5 x_{55},$$
$$y_1^{(2)} = 0.0590 + 0.2810 y_1^{(1)} + 0.3055 y_4^{(1)} + 0.3670 y_1^{(1)} y_4^{(1)},$$
$$y_2^{(2)} = 0.0225 + 0.4144 y_2^{(1)} + 0.3812 y_3^{(1)} + 0.1878 y_2^{(1)} y_3^{(1)},$$
$$y_3^{(2)} = 0.0609 + 0.2917 y_1^{(1)} + 0.2738 y_5^{(1)} + 0.3880 y_1^{(1)} y_5^{(1)},$$
$$y_1^{(3)} = 0.0551 + 0.3033 y_1^{(2)} + 0.3896 y_2^{(2)} + 0.2540 y_1^{(2)} y_2^{(2)},$$
$$y_2^{(3)} = 0.0579 + 0.4058 y_2^{(2)} + 0.2834 y_3^{(2)} + 0.2549 y_2^{(2)} y_3^{(2)},$$
$$y_1^{(4)} = -0.0400 + 0.6196 y_1^{(3)} + 0.5702 y_2^{(3)} - 0.1504 y_1^{(3)} y_2^{(3)},$$

where $x_5$ is absolute power of sub-delta in C4, $x_6$ is absolute power of sub-delta, $x_{21}$ is real power of alpha, $x_{28}$ is absolute power of beta1 in C3, $x_{55}$ is absolute variance of theta in C4, $x_{57}$ is absolute variance of sub-delta, and $x_{62}$ is absolute variance of sub-delta in C3. These variables were normalized.

Note that an extracted polynomial rule uses 7 from 72 input variables. Fig. 9 depicts an appropriate structure of the trained PNN, which consists of 7 input nodes and 11 neurons whose transfer function is described by (3)

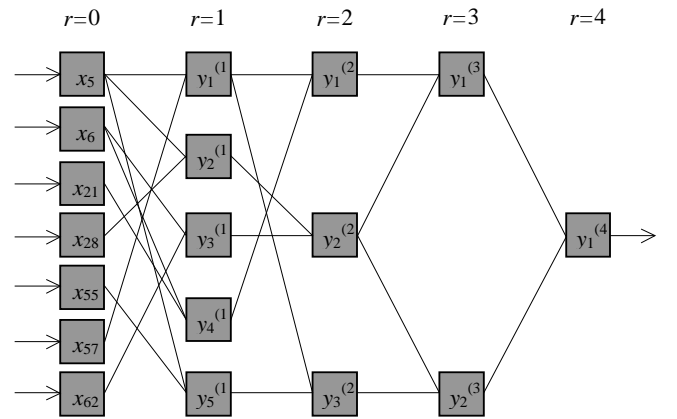

Fig. 9: A network for classifying normal and artifact EEG segments

Above polynomial rule may be used to automatically classify the EEG segments as follows:

```
If the output h ≥ 0.5, then an EEG segment
is normal, otherwise an artifact.
```

## VI. CONCLUSION

A common feature of fully connected feed-forward neural networks trained on real-world data is that the network structures must be well predefined. For this reason, the users have to experimentally search of the best structures of the neural networks that include minimal number of the input nodes and hidden neurons.

In contrast to the FNNs, the GMDH-type neural networks do not need to predefine their structures because the training algorithm grows an appropriate neural network. A resulting network is described by a concise set of short-term polynomials. However the GMDH-type training algorithms cannot effectively fit the neuron weights to real data distorted by a non-Gauss noise. As a result, the polynomial networks are often over-fitted.

A training algorithm we developed is able more effectively fit the neuron weights to the real data. The algorithm does not require to hypothesize a structure of the noise that distorts the training data.

In our experiments we used three types of real data: the EEG recordings from Alzheimer and healthy patients, the EEG data from the sudden death risk patients and the EEG recordings including the normal and artifact segments. These data, that medical experts manually cleared and labeled, we used to compare the classification accuracy of the trained FNNs and GMDH-type networks.

On the first EEG data, all three networks have the same accuracy. In more complex case, the polynomial classification rules that our algorithm extracted are performed slightly better than the standard FNNs and GMDH-type networks. The extracted rules correctly classify 72% of the risk group patients and 96.5% of the EEG segments on the testing datasets.

Thus we conclude that a neural network based technique we developed is able successfully to extract the polynomial classification rules from the EEG data. We hope that this technique can be also applied to other real-world problems.

The author is grateful to Frank Pasemann and Joachim Schult from TheoriLabor for fruitful discussion and also to Joachim Frenzel and Burghart Scheidt from Pediatric Clinic of the University Jena for their EEG recordings.

## REFERENCES

[1] C. Anderson, S. Devulapalli, E. Stolz. EEG Signal Classification with Different Signal Representations. In: F. Girosi, J. Makhoul, E. Manolakos, E. Wilson (eds.), Neural Networks for Signal Processing, IEEE Service Center, Piscataway, NJ, 475-483, 1995.

[2] M. Galicki, H. Witte, J. Dörschel, A. Doering, M. Eiselt, G. Grießbach. Common Optimization of Adaptive Preprocessing Units and a Neural Network During the Learning Period. Application in EEG Pattern Recognition. Neural Networks, 10:1153-1163, 1998.

[3] E. Riddington, E. Ifeachor, E. Allen, N. Hudson, D. Mapps. A Fuzzy Expert System for EEG Interpretation. In: E.C. Ifeachor, K.G. Rosén (eds.), Proc. of Conference on Neural Networks and Expert Systems in Medicine and Healthcare, NNESMED' 94, Plymouth, England, University of Plymouth, 291-302, 1994.

[4] O. Breidbach, K. Holthausen, B. Scheidt, J. Frenzel. Analysis of EEG Data Room in Sudden Infant Death Risk Patients. Theory Bioscience, 117:377-392, 1998.

[5] C. James, K. Kobayashi, J. Gotman. Seizure Detection With the Self-organizing feature map. In: H. Malmgren, M. Borga, L. Niklasson (eds.), Proc. of International Conference on Artificial Neural Network in Medicine and Biology, ANNIMAB'1, Goteborg, Sweden, May 2000, Springer-Verlag, 2000.

[6] S.J. Farlow (ed.). Self-organizing Methods in Modeling: GMDH Type Algorithms. Marcel Dekker Inc., New York and Basel, 1984.

[7] H.R. Madala, A.G. Ivakhnenko. Inductive Learning Algorithms for Complex Systems Modeling. CRC Press Inc., Boca Raton, 1994.

[8] J.-A. Müller, F. Lemke, A.G. Ivakhnenko. GMDH Algorithms for Complex Systems Modeling. Mathematical and Computer Modeling of Dynamical Systems, 4:275-315, 1998.

[9] H. Iba, T. Sato, H.. Genetic Programmin with Local Hill-Climbing. In: Davidir, Y., Schwefel H. P., Männer R. (eds.), Parallel Problem Solving from Nature, Springer-Verlag, 302-311, 1994.

[10] H. Iba, H. de Garis. Genetic Programming Using the Minimum Description Length Principle. In: K. Kinner (eds.), Advances in Genetic Programming, MIT Press, Cambridge, MA, 265-284, 1994.

[11] N.L. Nikolaev, H. Iba. Automated Discovery of Polynomials by Inductive Genetic Programming. In: J.M. Zutkow and J. Ranch (eds.), Principles of Data Mining and Knowledge Discovery, Third European Conference, PKDD'99, LNAI-1704, Springer, Berlin, 456-461, 1999.

[12] V.N. Fomin, A.L. Fradkov, V.A. Yakubovich. Adaptive Control of Dynamical Objects. Nauka, Moscow, 1981 (in Russian).

[13] URL: http://www.scri.fsu.edu/~nayak/chaos/data.html